\documentclass[runningheads]{llncs}
\usepackage[T1]{fontenc}
\usepackage[utf8]{inputenc}
\usepackage{graphicx,verbatim}
\usepackage{amsmath}
\usepackage{multirow}
\usepackage{graphicx}  
\usepackage{amsmath}   
\usepackage{amssymb}   
\usepackage{cite}      
\usepackage{algorithmic} 
\usepackage{array}     
\usepackage{url}       
\usepackage{hyperref}  
\usepackage{bm}
\usepackage{booktabs}  
\usepackage{caption}   
\usepackage{xcolor}      
\usepackage{colortbl}    
\usepackage{subcaption}  
\makeatletter
\def\@makefnmark{\hbox{\@textsuperscript{\@thefnmark}}}
\makeatother
\begin{document}
\title{DSA-SRGS: Super-Resolution Gaussian Splatting for Dynamic Sparse-View DSA Reconstruction}
\titlerunning{DSA-SRGS}

\author{%
Shiyu Zhang\inst{1}\protect\footnotemark[1] \and
Zhicong Wu\inst{2}\protect\footnotemark[1] \and
\setcounter{footnote}{3}
Huangxuan Zhao\inst{1}\protect\footnotemark[4] \and
Zhentao Liu\inst{3} \and
Lei Chen\inst{4} \and
Yong Luo\inst{1} \and
Lefei Zhang\inst{1} \and
Zhiming Cui\inst{3} \and
Ziwen Ke\inst{3}\protect\footnotemark[4] \and
Bo Du\inst{1}\protect\footnotemark[4]
}

\authorrunning{S. Zhang et al.}

\institute{%
School of Computer Science, Wuhan University, Wuhan, China \and
Institute of Artificial Intelligence, Xiamen University, Xiamen, China \and
School of Biomedical Engineering \& State Key Laboratory of Advanced Medical Materials and Devices, ShanghaiTech University, Shanghai, China \and
Union Hospital, Tongji Medical College, Huazhong University of Science and Technology, Wuhan, China
}

\maketitle              
\renewcommand{\thefootnote}{\fnsymbol{footnote}}
\footnotetext[1]{Equal Contribution.}
\footnotetext[4]{Corresponding authors.}
%
\begin{abstract}
Digital subtraction angiography (DSA) is a key imaging technique for the auxiliary diagnosis and treatment of cerebrovascular diseases. 
Recent advancements in gaussian splatting and dynamic neural representations have enabled robust 3D vessel reconstruction from sparse dynamic inputs.
However, these methods are fundamentally constrained by the resolution of input projections, where performing naive upsampling to enhance rendering resolution inevitably results in severe blurring and aliasing artifacts.
Such lack of super-resolution capability prevents the reconstructed 4D models from recovering fine-grained vascular details and intricate branching structures, which restricts their application in precision diagnosis and treatment. 
To solve this problem, this paper proposes DSA-SRGS, the first super-resolution gaussian splatting framework for dynamic sparse-view DSA reconstruction.
Specifically, we introduce a Multi-Fidelity Texture Learning Module that integrates high-quality priors from a fine-tuned DSA-specific super-resolution model, into the 4D reconstruction optimization.
To mitigate potential hallucination artifacts from pseudo-labels, this module employs a Confidence-Aware Strategy to adaptively weight supervision signals between the original low-resolution projections and the generated high-resolution pseudo-labels.
Furthermore, we develop Radiative Sub-Pixel Densification, an adaptive strategy that leverages gradient accumulation from high-resolution sub-pixel sampling to refine the 4D radiative gaussian kernels.
Extensive experiments on two clinical DSA datasets demonstrate that DSA-SRGS significantly outperforms state-of-the-art methods in both quantitative metrics and qualitative visual fidelity.

\keywords{Sparse-View DSA Reconstruction \and Gaussian Splatting  \and Super Resolution.}

\end{abstract}
\section{Introduction}  
Digital subtraction angiography (DSA)\cite{harrington1982digital} clearly presents vascular dynamics with high spatiotemporal resolution two-dimensional imaging, providing a key basis for the diagnosis and treatment of lesions\cite{liu2018digital}\cite{wang2019diagnostic}\cite{chen2020role}. 
The clear imaging of traditional DSA relies on dense projection acquisition, resulting in high radiation dose and easy introduction of motion artifacts. Therefore, achieving high-quality 4D DSA reconstruction under sparse-view conditions has become an important research direction in this field\cite{zhao2025large}\cite{zhao2026generative}\cite{xu2025garamost}\cite{xu2024most}.

\begin{figure}[t]
\centering
\includegraphics[width=0.8\textwidth]{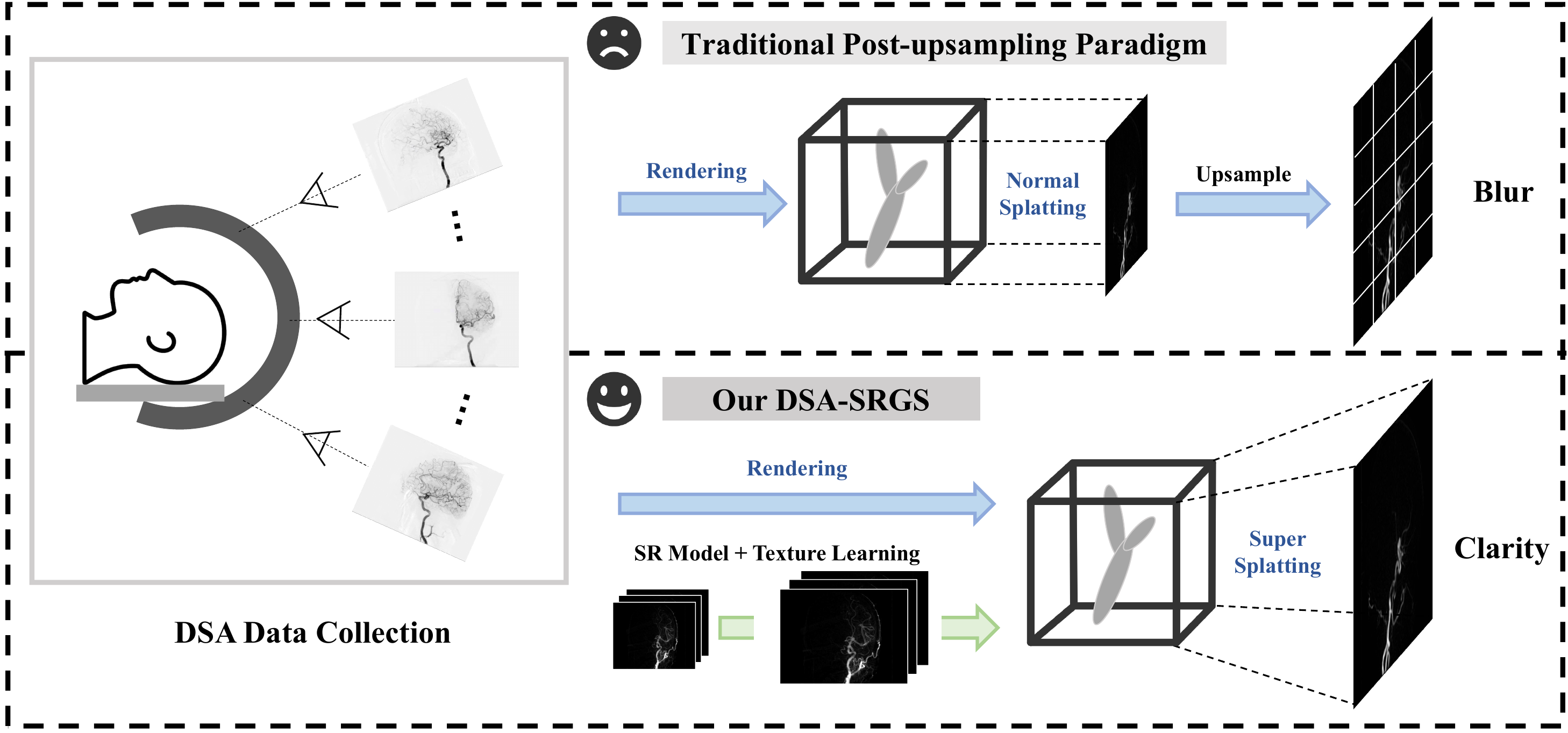}
\caption{Overview of traditional post-upsampling paradigm vs. our DSA-SRGS.} \label{first}
\end{figure}

Although some studies have successfully applied 3D gaussian splatting\cite{kerbl20233d} to the field of medical imaging\cite{zha2024rectifying} and introduced dynamic modeling and cumulative pruning strategies for DSA time series\cite{zhang2025togs}\cite{liu20254drgs}, the reconstruction quality is still limited by the resolution of the input projection. Under low-resolution conditions, the rendering results are prone to problems such as edge blurring, aliasing artifacts and noise sensitivity, which seriously restricts its clinical application in precision diagnosis and treatment\cite{el2023optimization}\cite{ruedinger20214d}\cite{sandoval20164d}\cite{lang20174d}\cite{sandoval2015comparison}.

It is worth noting that the mainstream computer vision method for solving the problem of limited image resolution is image super-resolution (SR) technology. SR research mainly focuses on network architecture innovation, covering CNN methods\cite{lecun2002gradient} sSRCNN\cite{dong2014learning}, EDSR\cite{lim2017enhanced}, and RCAN\cite{zhang2018image}. The performance has been continuously improved by Transformer models\cite{vaswani2017attention} such as SwinIR\cite{liang2021swinir}, DAT\cite{chen2023dat}, Mambair\cite{guo2025mambairv2}and HAT\cite{chen2023hat}, as well as GAN methods\cite{goodfellow2014generative} like Real-ESRGAN \cite{wang2021realesrgan}. In response to the demand for lightweighting, CATANet\cite{liu2025catanet} achieves efficient and long-range information interaction through content-aware token aggregation, and has achieved leading performance in lightweight super-resolution tasks.
While recent frontiers like SRGS \cite{feng2024srgs} have begun integrating SR priors into Gaussian splatting, their efficacy remains limited in medical contexts. 
Due to the significant domain gap, these general-purpose models often introduce "hallucination artifacts" that compromise the physical fidelity and clinical reliability of vascular structures.

To address resolution bottlenecks, this paper proposes DSA-SRGS, a super-resolution gaussian splatting framework for dynamic sparse-view DSA reconstruction that unifies super-resolution learning and 4D dynamic reconstruction in end-to-end optimization, as shown in Fig.~\ref{first}.
Specifically, our method first uses a fine-tuned super-resolution model to perform super-resolution on the low-resolution input image to restore the detailed information at the original resolution. Subsequently, a multi-fidelity learning is introduced, which absorbs the high-frequency texture information provided by the super-resolution model while maintaining the structural authenticity consistent with the original observations. To suppress the illusory textures that may occur in the super-resolution model, we further design a confidence-aware fusion mechanism to adaptively weight the multi-source supervised signals. During the rendering stage, we propose a radiation sub-pixel densification strategy. By accumulating gradients during the sampling process of high-resolution sub-pixels, we guide the gaussian kernel to adaptively split in texture-rich regions, thereby enhancing the model's ability to restore fine structures.

In summary, the main contributions of this article include:
\begin{enumerate}
    \item We introduce DSA-SRGS, the first super-resolution gaussian splatting framework for dynamic sparse-view DSA reconstruction.

    \item We propose the Multi-fidelity Texture Learning Module and Radiative Sub-pixel Densification Strategy to enhance micro-vascular modeling.


    \item Experiments on two clinical DSA datasets have shown that DSA-SRGS outperforms the current SOTA method.
\end{enumerate}

\begin{figure}[t]
\centering
\includegraphics[width=0.95\textwidth]{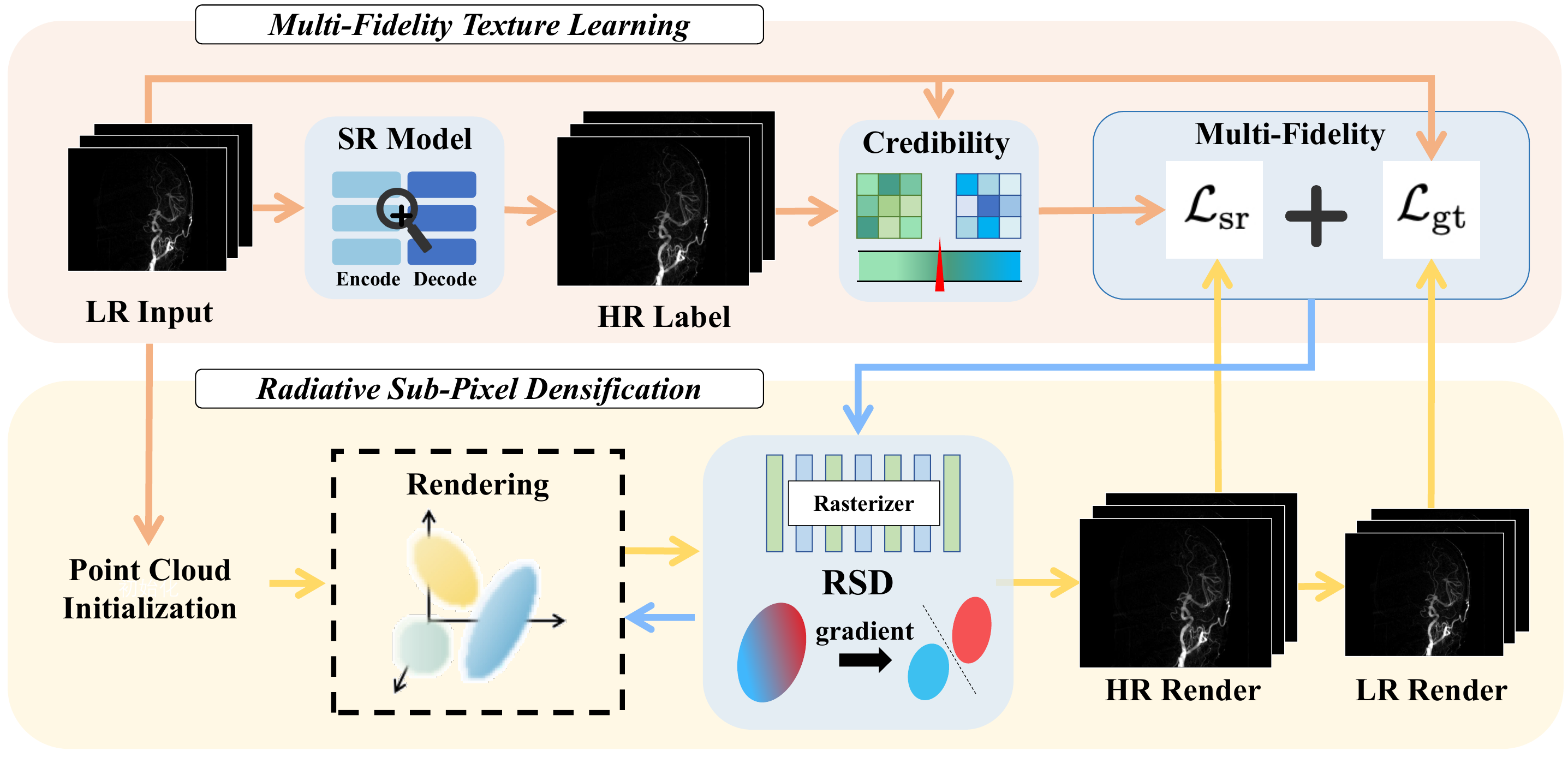}
\caption{Architecture of the proposed DSA-SRGS.} \label{second}
\end{figure}

\section{Method} 
\subsection{Problem Definition}
Given a low-resolution projection input $\mathcal{I}^{LR} = \{I_{t,v}^{LR} \in \mathbb{R}^{\frac{H}{4} \times \frac{W}{4}} \mid t = 1, \dots, T, v = 1, \dots, V\}$ of a dynamic DSA sequence, where \(T\) and \(V\) denote the total time frames and sparse viewpoints, respectively.
For each view \(v\), we construct the projection matrix \(P_v = K_v[R_v \mid t_v]\) based on its corresponding collected geometric parameters. 
DSA-SRGS aims to directly reconstruct a high-fidelity four-dimensional vascular model from these sparsely sampled low-resolution dynamic projections, and render high-resolution DSA images \(I^{HR} \in \mathbb{R}^{H \times W}\) from arbitrary viewpoints and at arbitrary time points.

\subsection{Radiative Sub-pixel Densification}
We represent vascular segments as gaussian kernels \(\mathcal{G} = \{g_i\}_{i=1}^{N}\), while introducing a time-dependent central attenuation function to capture the dynamic evolution of contrast agent concentration. To predict this attenuation value, we design a Dynamic Neural Attenuation Field (DNAF):
\begin{equation}
\rho_i(t) = \Phi\left(H_{3D}(\bm{\mu}_i) \oplus H_{4D}(\bm{\mu}_i, t)\right),
\end{equation}
where \(\mu_i\) denotes the position of the gaussian kernel, \(H_{3D}(\cdot)\)\cite{muller2022instant} and \(H_{4D}(\cdot,\cdot)\)\cite{park2023temporal} extract static scene features and spatio-temporal dynamic features respectively, and \(\Phi(\cdot)\) is an MLP that maps the features to the attenuation value \(\rho_i(t)\).

For rendering, We employ X-ray rasterization pipelines and cumulative attenuation pruning strategies:
\begin{equation}
\bar{\rho}_i = \frac{1}{N_{\text{iter}}} \sum_{k=1}^{N_{\text{iter}}} \rho_i(t_k),
\end{equation}
where $N_{\text{iter}}$ is the number of iterations within a pruning interval, and $t_k$ is the timestamp at the $k$-th iteration. 

Although pruning can clear the background, it cannot solve the problem of detail loss caused by insufficient resolution. To this end, inspired by Pixel-GS et al.\cite{zhang2024pixel}\cite{bulo2024revising}, this paper proposes the radiative sub-pixel densification strategy. By accumulating gradients in high-resolution sub-pixel sampling, it guides the gaussian kernel to adaptively split in texture-rich regions, thereby enhancing the ability to restore small vascular branches.

This strategy dynamically identifies regions requiring refinement based on the projection gradients of gaussian kernels. Regions with larger gradients require higher gaussian kernel density. Therefore, the densification condition is formulated as:
\begin{equation}
\mathcal{D}_{\text{RSD}} = \{g_i \mid \|\nabla_i\| > \tau_{\text{grad}} \cdot \eta_{\text{iter}}\},
\end{equation}
where \(g_i\) is the gaussian kernel, \(\nabla_i\) is the accumulated gradient, \(\tau_{\text{grad}}\) is the gradient threshold and \(\eta_{\text{iter}}\) is an iteration-dependent decay coefficient.

For each seed kernel \(g_i\) that meets the condition, RSD generates \(K\) sub-kernels, and the newly generated kernels form a finer overlay around the parent kernel:
\begin{equation}
\boldsymbol{\mu}_{i}^{(k)} = \boldsymbol{\mu}_i + \Delta\boldsymbol{\mu}_{i}^{(k)}, \quad \Delta\boldsymbol{\mu}_{i}^{(k)} \sim \mathcal{N}(0, \boldsymbol{\Sigma}_i \cdot \alpha),
\end{equation}
\begin{equation}
\boldsymbol{s}_{i}^{(k)} = \beta \cdot \boldsymbol{s}_i, \quad \beta \in (0,1),
\end{equation}
where \(\alpha\) controls the position offset magnitude and \(\beta\) is the scale decay factor.

Furthermore, this strategy introduces a residual-guided mechanism that actively adds small-scale gaussian kernels in high-residual regions using the difference map between rendered and ground-truth images.
\subsection{Multi-fidelity Texture Learning}
Although sub-pixel constraints promote gaussian kernel densification, the input low-resolution image lacks high-frequency textures, and it is difficult to restore the fine structure of blood vessels solely by internal constraints.
To provide external priors for 4D reconstruction, we introduce a dedicated super-resolution (SR) model for DSA to generate high-resolution textures. To mitigate the risk of the model learning unreliable illusory artifacts, we further design a multi-fidelity learning strategy. Specifically, a confidence-aware teaching reference image is constructed, which preserves SR-enhanced textures in high-confidence regions while reverting to the original upsampled observations in low-confidence areas:
\begin{equation}
C = \sigma\left(\alpha \cdot \mathcal{S}(I_{\text{SR}}, I_{\text{LR}}^{\uparrow}) + \beta \cdot \mathcal{T}(I_{\text{SR}})\right),
\end{equation}
\begin{equation}
I_{\text{teach}} = C \odot I_{\text{SR}} + (1 - C) \odot I_{\text{LR}}^{\uparrow},
\end{equation}
where $I_{\text{SR}}$ and $I_{\text{LR}}^{\uparrow}$ denote the SR-generated and upsampled LR images respectively; $\mathcal{S}(\cdot,\cdot)$ represents SSIM for local consistency; $\mathcal{T}(\cdot)$ is a texture richness assessment function; $\alpha, \beta$ are learnable scaling parameters; $\sigma(\cdot)$ denotes the Sigmoid function; and $\odot$ denotes element-wise multiplication.

The total loss is formulated as a weighted combination of a high-fidelity loss $\mathcal{L}_{\text{gt}}$ and a low-fidelity loss $\mathcal{L}_{\text{sr}}$. Specifically, $\mathcal{L}_{\text{gt}}$ is computed in the low-resolution space to ensure strict adherence to the original observations, while $\mathcal{L}_{\text{sr}}$ provides supervision in the high-resolution space via the teaching reference image:
\begin{equation}
\left\{
\begin{aligned}
\mathcal{L}_{\text{gt}} &= (1 - \lambda_{\text{ssim}}) \| I_{\text{rend}}^{\downarrow} - I_{\text{LR}} \|_1 + \lambda_{\text{ssim}} (1 - \text{SSIM}(I_{\text{rend}}^{\downarrow}, I_{\text{LR}})) \\[1.2ex]
\mathcal{L}_{\text{sr}} &= (1 - \lambda_{\text{ssim}}) \| I_{\text{rend}} - I_{\text{teach}} \|_1 + \lambda_{\text{ssim}} (1 - \text{SSIM}(I_{\text{rend}}, I_{\text{teach}}))
\end{aligned}
\right.,
\end{equation}
where $I_{\text{rend}}$ and $I_{\text{rend}}^{\downarrow}$ denote the rendered high-resolution projection and its downsampled low-resolution version, respectively; $I_{\text{LR}}$ is the original input projection; $I_{\text{teach}}$ represents the confidence-aware teaching reference; $\lambda_{\text{ssim}}$ is the balancing weight between the $\ell_1$ distance and the structural similarity index.

\section{Experiments} 
\subsection{Experiment Setup}
\subsubsection{Datasets}We evaluated the model performance on two clinical DSA datasets: DSA-15 was derived from the 4DRGS work (15 cases), and DSA-28 was self-collected data (28 cases), covering multi-center samples from the left, right, and back of the brain. In addition, an extra 135 clinical dynamic DSA sequences were collected. After digital subtraction processing, 17,822 vascular subtraction images were obtained for fine-tuning the super-resolution model.

\begin{table}[t]
  \centering
  \caption{Comparison with other reconstruction methods.  Bold indicates the best results. Underline indicates the second-best result.}
  \label{tab:1}
  \setlength{\tabcolsep}{8pt}    
    \begin{tabular}{cc|cc|cc}
    \toprule
    \multirow{2}{*}{View} & \multirow{2}{*}{Method} & \multicolumn{2}{c|}{DSA-28} & \multicolumn{2}{c}{DSA-15} \\
\cmidrule{3-6}          &       & PSNR $\uparrow$ & SSIM $\uparrow$ & PSNR $\uparrow$ & SSIM $\uparrow$ \\
    \midrule
    \multirow{4}{*}{30} & R$^2$-Gaussian & 28.163  & 0.7895  & 28.499  & 0.7871  \\
          & TOGS  & 33.338  & 0.8362  & 33.273  & 0.8349  \\
          & 4DRGS & \underline{33.819}  & \underline{0.8541}  & \underline{33.687}  & \underline{0.8433}  \\
          & Ours  & \textbf{34.323} & \textbf{0.8563} & \textbf{34.198} & \textbf{0.8543} \\
    \midrule
    \multirow{4}{*}{40} & R$^2$-Gaussian & 28.394  & 0.7947  & 28.716  & 0.7934  \\
          & TOGS  & 33.448  & 0.8384  & 33.417  & 0.8377  \\
          & 4DRGS & \underline{34.129}  & \underline{0.8568}  & \underline{34.017}  & \underline{0.8472}  \\
          & Ours  & \textbf{34.742} & \textbf{0.8600} & \textbf{34.645} & \textbf{0.8587} \\
    \bottomrule
    \end{tabular}%
\end{table}

\begin{figure}[t]
\centering
\includegraphics[width=0.8\textwidth]{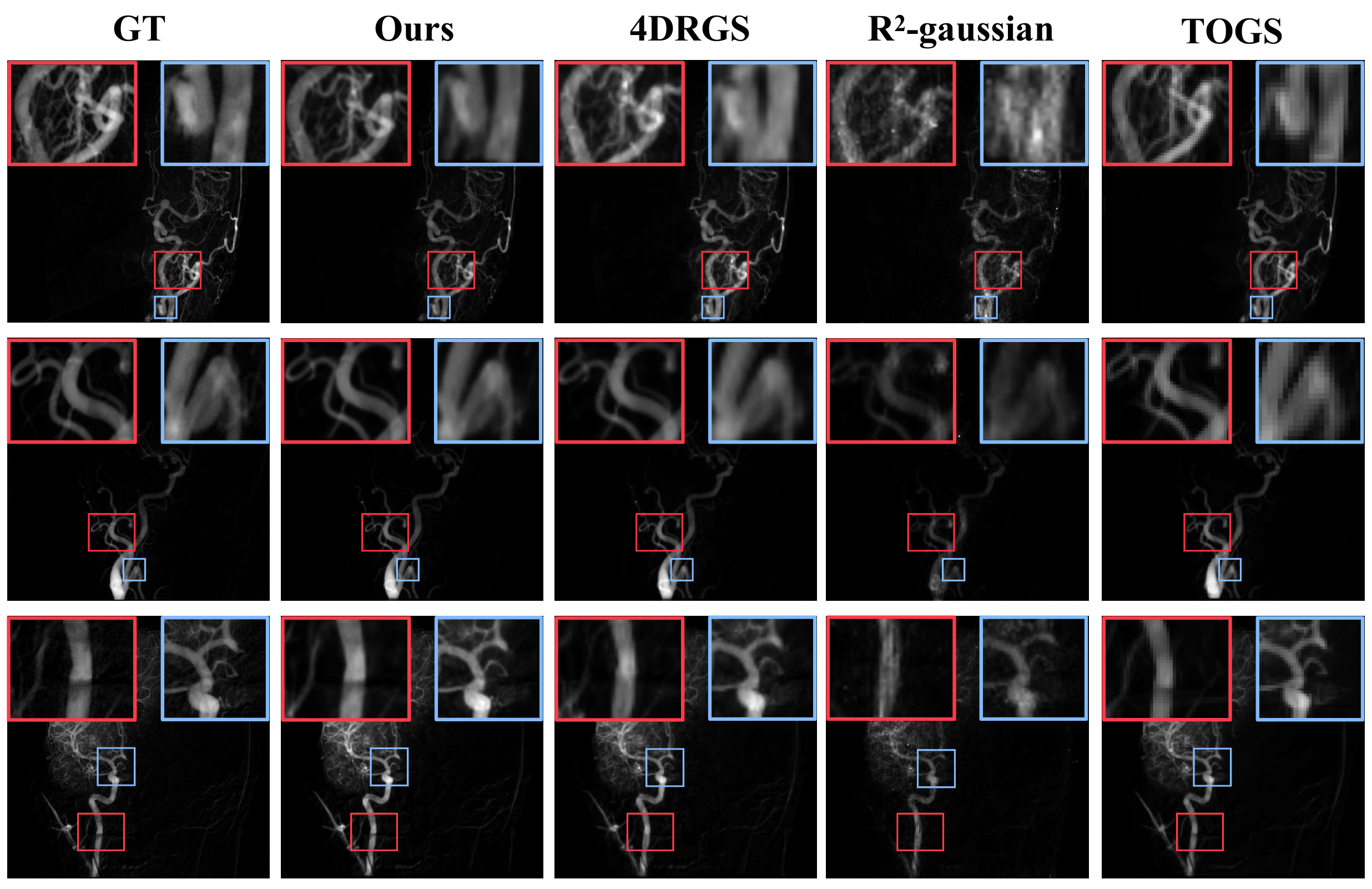}
\caption{Qualitative comparison of different models..} \label{third}
\end{figure}

\subsubsection{Evaluation Criteria}We use three complementary metrics to evaluate the generation quality of DSA images using three complementary metrics: pixel-level PSNR for reconstruction quality, patch-level SSIM\cite{wang2004image} for structural consistency.

\subsubsection{Implementation Details}For radiative sub-pixel densification, we set a gradient accumulation window of 100 iterations, using the same splitting threshold as the density control. Pruning follows the accumulated attenuation strategy with threshold \(\epsilon=1\times10^{-6}\). The initial number of gaussian kernels is \(M=30k\) with threshold \(\delta=0.016\). The multi-fidelity learning weight \(\beta=0.4\) and SSIM loss weight \(\lambda_{\text{ssim}}=0.2\). Experiments are conducted on a single RTX 3090 GPU, with each case reconstructed in approximately 15 minutes.

\subsection{Comparison With State-of-The-Art}
To evaluate DSA-SRGS, we conducted quantitative comparisons against R$^2$-Gaussian, TOGS, and 4DRGS on two clinical datasets. 

\subsubsection{Quantitative analysis}Table~\ref{tab:1} show that DSA-SRGS achieves optimal performance under all settings. Compared with the existing methods, this method achieves a PSNR of 34.32dB under 30 views and an LPIPS as low as 0.147, significantly enhancing the visual realism. From 30 to 40 views, the PSNR improvement of DSA-SRGS is superior to that of 4DRGS, demonstrating stronger sparse sampling robustness. On the two datasets, the performance of DSA-SRGS was consistent, and the SSIM was all above 0.85, verifying its effectiveness and generalization ability in clinical scenarios.

\subsubsection{Qualitative analysis}Fig.~\ref{third}) shows the comparison of DSA projections of different methods from 30 perspectives. The R$^2$-Gaussian structure is missing and artifacts are severe. The TOGS contour is rough and shows a distinct mosaic effect when magnified. The structure of 4DRGS is complete, but the details of the small branches are lost. In contrast, DSA-SRGS is significantly superior to the comparison methods, restoring the vascular margins and fine structures.

\subsection{Ablation Analysis}

We validate key components and super-resolution model effectiveness via ablation studies on clinical datasets as shown in table~\ref{tab:ablation_study_final} and ~\ref{tab:sr_model_comparison}.

\begin{table}[t]
\centering
\caption{Ablation Study of Core Components for DSA-SRGS. SR: SR Rendering, BS: Blurry Supervision, PL: Pseudo-label, MF: Multi-fidelity Learning, C: Confidence.}
\label{tab:ablation_study_final}
\setlength{\tabcolsep}{5pt}
\footnotesize
\begin{tabular}{cccccc|cc|cc}
\toprule
\multicolumn{6}{c|}{Components} & \multicolumn{2}{c|}{DSA-28} & \multicolumn{2}{c}{DSA-15} \\
\cmidrule{1-10}
Baseline & SR & BS & PL & MF & C & PSNR $\uparrow$ & SSIM $\uparrow$ & PSNR $\uparrow$ & SSIM $\uparrow$ \\
\midrule
$\checkmark$ & & & & & & 33.163 & 0.8405 & 33.819 & 0.8541 \\
$\checkmark$ & $\checkmark$ & $\checkmark$ & & & & 33.243 & 0.8417 & 34.030 & 0.8543 \\
$\checkmark$ & $\checkmark$ & & $\checkmark$ & & & 33.533 & 0.8312 & 34.144 & 0.8214 \\
$\checkmark$ & $\checkmark$ & $\checkmark$ & $\checkmark$ & & & 33.572 & 0.8493 & 34.223 & 0.8537 \\
$\checkmark$ & $\checkmark$ & & $\checkmark$ & $\checkmark$ & & \underline{33.673} & \underline{0.8504} & \underline{34.287} & \underline{0.8544} \\
$\checkmark$ & $\checkmark$ & & $\checkmark$ & $\checkmark$ & $\checkmark$ & \textbf{34.198} & \textbf{0.8543} & \textbf{34.323} & \textbf{0.8563} \\
\bottomrule
\end{tabular}
\end{table}


\subsubsection{Effect of Core Components} Table~\ref{tab:ablation_study_final} validates the incremental contribution of each component. Starting from the 4DRGS baseline (33.819 dB PSNR on DSA-15), Blurry Supervision (+BS) yields a marginal gain (+0.211 dB). While Pseudo-Labels (+PL) boost PSNR to 34.144 dB, they induce a notable SSIM degradation (-0.0327), signaling structural hallucinations. Integrating BS mitigates this distortion (+PL+BS: SSIM 0.8537). The Multi-Fidelity module (+MF) further optimizes the trade-off between high-frequency details and observational constraints. Ultimately, the full model (+MF+C) achieves peak performance (34.323 dB/0.8563), demonstrating that confidence-aware fusion effectively suppresses artifacts while Radiative Sub-pixel Densification enhances micro-vascular modeling. Consistent trends on DSA-28 verify generalization.

\begin{table}[t]
\centering
\caption{Quantitative comparison of super-resolution models.}
\label{tab:sr_model_comparison}
\setlength{\tabcolsep}{10pt}    
\small
\begin{tabular}{l|cc|cc}
\toprule
\multirow{2}{*}{SR Model} & \multicolumn{2}{c|}{DSA-28} & \multicolumn{2}{c}{DSA-15} \\
\cmidrule(lr){2-3} \cmidrule(lr){4-5}
 & PSNR $\uparrow$ & SSIM $\uparrow$ & PSNR $\uparrow$ & SSIM $\uparrow$ \\
\midrule
SwinIR\cite{liang2021swinir} & 33.547 & 0.8479 & 34.190 & 0.8536 \\
Real-ESRGAN\cite{wang2021realesrgan} & 32.988 & 0.8430 & 33.601 & 0.8468 \\
EDT\cite{li2021efficient} & 33.536 & 0.8475 & 34.176 & 0.8526 \\
HAT\cite{chen2023hat} & 33.526 & 0.8485 & 34.164 & 0.8532 \\
Mambair\cite{guo2025mambairv2} & 33.621 & 0.8499 & 34.253 & 0.8554 \\
CATANet\cite{liu2025catanet} & \underline{33.805} & \underline{0.8518} & \underline{34.257} & \underline{0.8554} \\
\textbf{Ours} & \textbf{34.198} & \textbf{0.8543} & \textbf{34.323} & \textbf{0.8563} \\
\bottomrule
\end{tabular}
\end{table}




\subsubsection{Effect of Super-resolution Model} To establish the optimal pseudo-label generator, we benchmarked mainstream super-resolution architectures across both clinical datasets as shown in table~\ref{tab:sr_model_comparison}. 
CATANet emerged as the superior base model, exhibiting robust cross-center generalization with minimal performance variance, attributed to its content-aware token aggregation mechanism that effectively models vascular dynamics. 
%
Crucially, our domain-specific fine-tuning yielded consistent gains across both clinical datasets, surpassing the pre-trained CATANet by 0.066/0.393 dB in PSNR and 0.0009/0.0025 in SSIM, respectively. This simultaneous improvement in fidelity and structural similarity underscores the necessity of adapting general SR priors to the DSA domain. 

\section{Conclusion} 
This paper proposes DSA-SRGS, a super-resolution gaussian splatting framework for DSA reconstruction of dynamic sparse views. This method integrates the prior of the super-resolution model into 4D reconstruction through a multi-fidelity texture learning module, and uses a confidence-aware strategy to balance multi-source supervision and suppress illusion artifacts. At the same time, radiative sub-pixel densification is introduced to guide the gaussian kernel to adaptively split in texture-rich areas, enhancing the modeling ability for small blood vessels. Experiments show that DSA-SRGS outperforms existing methods in both quantitative indicators and visual quality. Future work will explore self-supervised super-resolution reconstruction and further optimize the reconstruction speed to expand its clinical application.
\bibliographystyle{splncs04}  
\bibliography{DSA-SRGS_arxiv}     
\end{document}